\documentclass[10pt,twocolumn,letterpaper]{article}

\usepackage[pagenumbers]{cvpr}

\usepackage{multirow}
\usepackage{tabularx,array,makecell,pifont}
\usepackage{cuted}

\newcommand{\expfigure}[3]{%
  \IfFileExists{#1}{%
    \includegraphics[width=\linewidth]{#1}%
  }{%
    \fbox{\parbox[c][#2][c]{\dimexpr\linewidth-2\fboxsep-2\fboxrule\relax}{%
      \centering\footnotesize #3}}%
  }%
}
\providecommand{\stagyes}{\ding{51}}
\providecommand{\stagno}{\ding{55}}

\definecolor{cvprblue}{rgb}{0.21,0.49,0.74}
\usepackage[pdfversion=1.7,breaklinks,colorlinks,allcolors=cvprblue]{hyperref}

\def\confName{CVPR}
\def\confYear{2027}

\title{HuScale: Scalable Task-Level Data Generation for Humanoid Loco-Manipulation with KeypointVLA}

\author{%
  Guanyu Cai\textsuperscript{1,*}\quad
  Zheng Cao\textsuperscript{2,*}\quad
  Quanbao Lin\textsuperscript{1,*}\quad
  Tengjie Zhu\textsuperscript{1}\\[3pt]
  Guanzhu Ren\textsuperscript{1}\quad
  Yu Zhang\textsuperscript{1}\quad
  Tianchen Deng\textsuperscript{1,\textdagger}\quad
  Yao Mu\textsuperscript{1,\textdagger}\\[5pt]
  \normalsize\textsuperscript{1}Shanghai Jiao Tong University
  \qquad \textsuperscript{2}Zhejiang University\\[3pt]
  \small\textsuperscript{*}Equal contribution.
  \qquad \textsuperscript{\textdagger}Corresponding authors.
  \\[4pt]\small Project page\enspace\url{https://guanyucai-sjtu.github.io/huscale-project-page/}
}
\hypersetup{pdftitle={HuScale: Scalable Task-Level Data Generation for Humanoid Loco-Manipulation with KeypointVLA},pdfauthor={Guanyu Cai, Zheng Cao, Quanbao Lin, Tengjie Zhu, Guanzhu Ren, Yu Zhang, Tianchen Deng, Yao Mu}}

\begin{document}
\maketitle
\begin{strip}
    \centering
    \includegraphics[width=\linewidth]
        {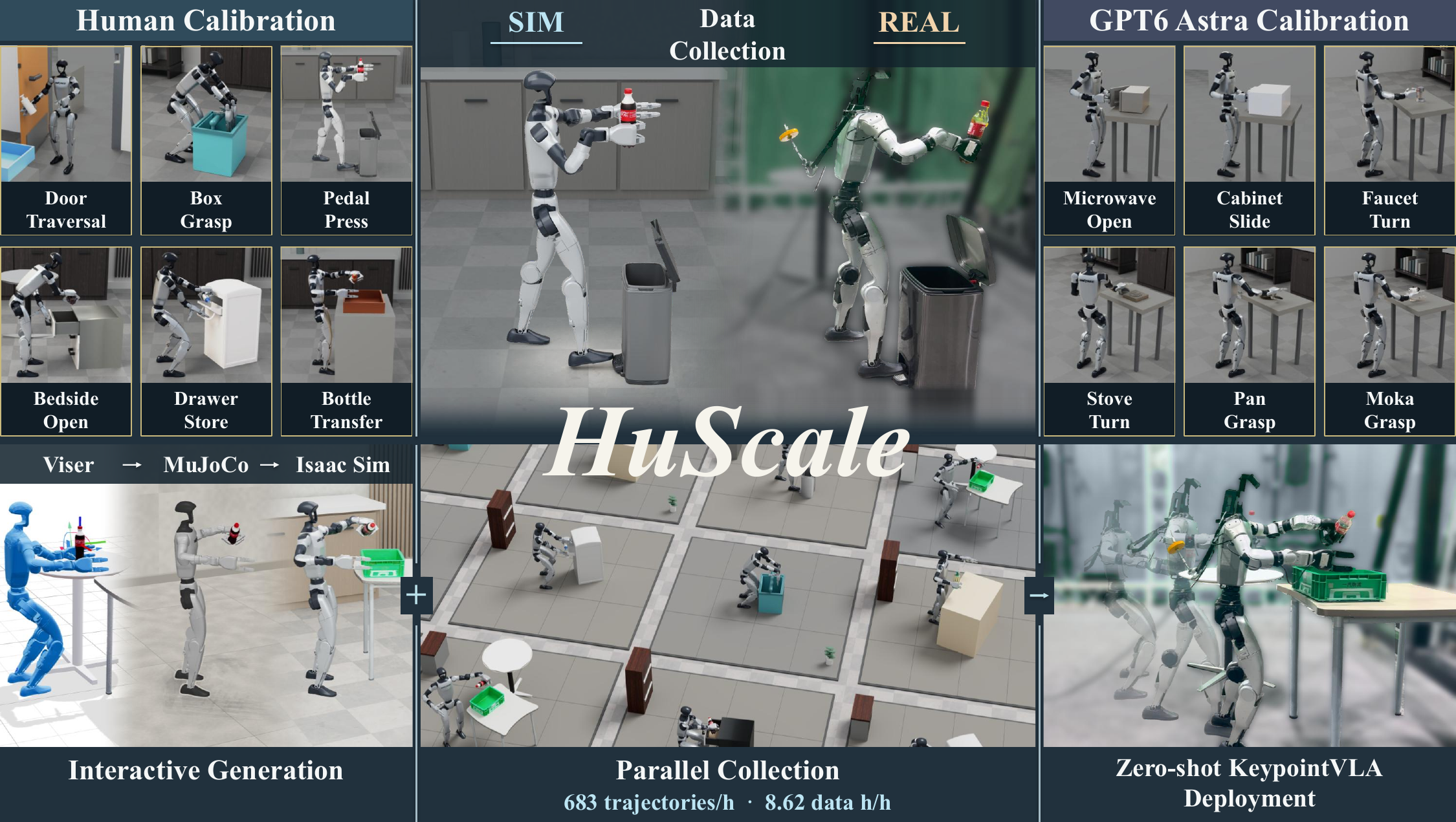}

    \captionsetup{hypcap=false}
    \captionof{figure}{
    \textbf{HuScale: scalable task-level data generation for humanoid loco-manipulation with KeypointVLA.}
    \textbf{Top:} Simulated/real Pedal Disposal, flanked by human-interactive
    and GPT6 Astra calibration examples.
    \textbf{Bottom:} Calibration-to-execution workflow, parallel collection,
    and zero-shot real-robot deployment.
    }
    \label{fig:teaser}
    \captionsetup{hypcap=true}
\end{strip}

\begin{abstract}
Scaling humanoid loco-manipulation requires diverse demonstrations
of whole-body interaction.
However, reliance on human motion demonstrations or per-trajectory engineering remains a major bottleneck to scaling up data generation.
We introduce \textbf{HuScale}, a {pipeline} for scalable task-level
data generation and humanoid loco-manipulation with
\textbf{KeypointVLA}.
{For data generation, reusable atomic skills are calibrated
through object-centric keypoint poses specified at keyframes by human
operators or GPT6 Astra.}
{HuScale then composes these skills into whole-body reference
sequences.}
{A shared whole-body tracker is trained to execute these
references across randomized scenes, both in simulation and in the real world, enabling scalable task-level data
generation without task-specific human motion demonstrations.}
{Successful rollouts yield
synchronized visual observations and whole-body actions paired with task
instructions.}
{For policy learning and execution, KeypointVLA uses a
keypoint-based action head to learn spatial targets from the generated
demonstrations. Conditioned on previous action chunks, it learns through
blended-target supervision to support
error correction and continuity across chunks.}
{Experiments show that HuScale generates up to 683 trajectories
(8.62 hours of demonstration data) per hour on a single NVIDIA
RTX~4090.}
{A data study across three VLA backbones and six tasks
verifies the utility of the generated demonstrations for policy training.}
{We also demonstrate that KeypointVLA, combined with visual
augmentation, enables zero-shot transfer to a physical Unitree G1 without
real-world demonstrations.}
\end{abstract}

\section{Introduction}

Humanoid robots are well suited to physical tasks in human environments.
Such tasks require coordinated locomotion and whole-body manipulation to
interact with {rigid and articulated objects}.
Recent advances in vision--language--action (VLA) models~\cite{kim2025openvla,bjorck2025grootn1,hu2026openhlm} and whole-body
control~\cite{peng2021amp,luo2023phc,he2025hover,luo2026sonic,zhu2026clot,yang2026sphlm} have enabled increasingly capable humanoid
loco-manipulation.
{However, training VLA policies for diverse humanoid
loco-manipulation tasks remains constrained by the availability of suitable
demonstrations.}
Unlike fixed-base manipulation, humanoid demonstrations must capture
locomotion, balance, contact, and whole-body coordination across broad
spatial and temporal scales.
{For VLA training, these demonstrations further include visual
observations synchronized with whole-body actions, along with task instructions.}
{This motivates generating diverse, dynamically feasible
demonstrations at scale.}

Recent work has explored motion retargeting~\cite{yang2025omniretarget},
video-based reconstruction~\cite{xie2026grail}, demonstration
adaptation~\cite{mandlekar2023mimicgen,jiang2025dexmimicgen,garrett2025skillmimicgen,lin2026humanoidmimicgen}, and learned or procedural data
generation~\cite{kim2026legs,wang2026vlk,rayyan2026fetchman,ling2026gentrack}.
Although these approaches substantially reduce direct robot data
collection, they rarely address task-level scalability, demonstration-level
dynamic feasibility, reusable execution, and learning-ready whole-body
supervision together.
{More broadly,} we ask whether \emph{the task itself, rather than an individual
trajectory, can become the unit of humanoid data scaling}.

We introduce \textbf{HuScale} (Fig.~\ref{fig:teaser}), {a pipeline for
scalable task-level data generation and humanoid loco-manipulation with
\textbf{KeypointVLA}.
Human operators or GPT6 Astra calibrate reusable atomic skills through
object-centric keypoint poses at keyframes.
These skills compose into tasks instantiated across randomized scenes
without per-trajectory editing.
For locomotion, a pretrained SONIC~\cite{luo2026sonic} locomotion prior converts A* routes into
whole-body references; for manipulation, online whole-body
inverse kinematics generates references from skill targets and measured states.
A trained shared whole-body tracker executes these references both in simulation
and on hardware. Successful rollouts provide synchronized visual observations
and whole-body actions paired with task instructions, alongside robot states
and control traces.}

To learn from these demonstrations, we develop \textbf{KeypointVLA},
{whose keypoint-based action head predicts body-keypoint poses
and wrist/finger-joint targets.
Previous chunk injection conditions training on previous action chunks.
Blended-target supervision supports corrective updates, while the temporal loss promotes
continuity across action chunks.
The shared tracker executes these targets with live-state
feedback, coupling visual replanning with whole-body control for multi-stage
loco-manipulation involving articulated-object and foot manipulation.}

{Using HuScale, we collect \textbf{12,000 demonstrations across six tasks
(2,000 per task)} and evaluate them across three VLA backbones. Compatible trajectories are also reused with SONIC.
With visual augmentation, KeypointVLA achieves zero-shot transfer to a physical
Unitree G1 without real-world demonstrations or fine-tuning.
Our contributions are threefold:}
\begin{itemize}
    \item \textbf{Task-level scalable and dynamically feasible data generation.}
    HuScale enables task-level data scaling by composing reusable atomic
    skills into tasks and instantiating them across randomized scenes.
    Closed-loop execution of spatial references {in simulation and on hardware} produces {synchronized}
    vision--language--whole-body-action demonstrations.

    \item {\textbf{KeypointVLA with a keypoint-based action head.}
    KeypointVLA predicts explicit body-keypoint targets across VLA backbones.
    Previous chunk injection, blended-target supervision, and a temporal
    loss support error correction and continuity.
    The shared tracker follows the predicted targets for closed-loop control.}

    \item \textbf{Systematic evaluation of data utility and transfer.}
    {We evaluate generation efficiency, data utility across multiple
    VLA backbones, demonstration reuse through alternative
    whole-body execution interfaces, and zero-shot sim-to-real transfer of
    KeypointVLA.}
\end{itemize}

\section{Related Works}
\label{sec:related_work}


\begin{table*}[!t]

\centering
\footnotesize
\setlength{\tabcolsep}{2pt}
\renewcommand{\arraystretch}{1.15}

\caption{\textbf{Comparison with recent humanoid data-generation and loco-manipulation methods.}
\stagyes{}: reported support; \stagno{}: no reported support.}
\label{tab:generation_comparison}

\begin{tabularx}{\textwidth}{@{}
    >{\raggedright\arraybackslash}p{0.19\textwidth}
    *{7}{>{\centering\arraybackslash}X}
    @{}}
\toprule
& \multicolumn{3}{c}{\textbf{Data generation}}
& \multicolumn{2}{c}{\textbf{Policy and execution}}
& \multicolumn{2}{c}{\textbf{Interaction capability}} \\
\cmidrule(lr){2-4}
\cmidrule(lr){5-6}
\cmidrule(l){7-8}

\textbf{Method}
& \makecell[c]{No human\\interaction\\demos}
& \makecell[c]{{Task}\\{Composition}}
& \makecell[c]{{Automated}\\{real-robot}\\{collection}}
& \makecell[c]{Body\\ref./latent}
& \makecell[c]{Shared\\executor}
& \makecell[c]{{Articulated-}\\{object}\\{manipulation}}
& \makecell[c]{{Foot}\\{manipulation}} \\
\midrule

DemoHLM~\cite{fu2025demohlm} & \stagno & \stagno & \stagno & \stagno & \stagyes & \stagyes & \stagno \\

\makecell[l]{HumanoidMimicGen~\cite{lin2026humanoidmimicgen}} & \stagno & \stagno & \stagno & \stagno & \stagyes & \stagyes & \stagno \\

LEGS~\cite{kim2026legs} & \stagyes & \stagyes & \stagno & \stagno & \stagyes & \stagno & \stagno \\

GRAIL~\cite{xie2026grail} & \stagyes & {\stagyes} & \stagno & \stagyes & \stagno & \stagno & \stagno \\

VLK~\cite{wang2026vlk} & \stagno & {\stagyes} & \stagno & \stagyes & \stagyes & \stagno & \stagno \\

{FetchMan~\cite{rayyan2026fetchman}} & {\stagyes} & {\stagno} & {\stagno} & {\stagno} & {\stagyes} & {\stagno} & {\stagno} \\

Zero-HLM~\cite{yang2026zerohlm} & \stagno & \stagno & \stagyes & \stagyes & \stagyes & \stagyes & \stagyes \\

\midrule
\textbf{HuScale (ours)} & \stagyes & \stagyes & \stagyes & \stagyes & \stagyes & \stagyes & \stagyes \\
\bottomrule
\end{tabularx}
\vspace{-2pt}
\end{table*}

\paragraph{Humanoid data collection and generation.}
HumanPlus~\cite{fu2024humanplus} uses camera-based shadowing,
while OmniH2O~\cite{he2024omnih2o} and TWIST2~\cite{ze2025twist2} support VR teleoperation.
For generation, DemoHLM~\cite{fu2025demohlm} adapts one demonstration through object-relative
transformations, and HumanoidMimicGen~\cite{lin2026humanoidmimicgen} combines
demonstration adaptation with whole-body planning.
Zero-HLM~\cite{yang2026zerohlm} combines scene reconstruction with demonstration-anchored
motion synthesis.
GRAIL~\cite{xie2026grail} uses generated interaction videos and task-family-specific
tracking adaptors.
VLK~\cite{wang2026vlk} combines motion priors with an interaction generator trained on
retargeted demonstrations.
LEGS~\cite{kim2026legs} composes motion primitives and renders observations with
Gaussian splatting.
{FetchMan~\cite{rayyan2026fetchman} uses scripted control with privileged state},
while SIMPLE~\cite{wei2026simple} combines arm planning with scripted locomotion.
On the rendering side, neural scene representations support novel-view
synthesis~\cite{deng2025best3dscenerepresentation}, and Gaussian SLAM methods
such as MG-SLAM~\cite{liu2025mgslam} reconstruct indoor scenes from RGB-D observations.

{Table~\ref{tab:generation_comparison} compares data generation,
policy and execution, and interaction capabilities.
HuScale composes atomic skills calibrated by humans or GPT6 Astra without
requiring human interaction demonstrations.
Once calibrated, these reusable atomic skills support scene-conditioned
collection without continuous human motion input.
It combines articulated-object and foot manipulation with a shared
whole-body tracker for simulation and automated real-robot collection.}

\paragraph{High-level VLA training.}
GR00T N1~\cite{bjorck2025grootn1} jointly trains vision--language and action modules
on real-robot, human-video, and synthetic data.
$\psi_0$~\cite{wei2026psi0} separates egocentric-video pretraining from humanoid
action-expert training.
OpenHLM~\cite{hu2026openhlm} adapts a manipulation-pretrained VLA to whole-body
joint commands and co-trains on stationary interaction data.
Workhorse~\cite{hu2026workhorse} learns five-link pose plans from human demonstrations
with augmentations addressing planner--tracker mismatch.
KeypointVLA pairs frozen vision--language backbones with a common action-head
architecture for six-keypoint trajectories (Sec.~\ref{sec:spatial_policy}).
Training combines blended-target supervision in physical units with
previous chunk injection inspired by self-forcing~\cite{huang2025selfforcing}.

\paragraph{Low-level whole-body tracker training.}
OmniH2O~\cite{he2024omnih2o} distills a privileged reinforcement-learning teacher into a
deployable tracker conditioned on sparse motion goals and
proprioceptive history.
SONIC~\cite{luo2026sonic} learns a shared motion-token space through PPO~\cite{schulman2017ppo} with reconstruction
and alignment objectives.
SuperMimic~\cite{liao2026supermimic} fits differentiable dynamics to simulator transitions and
improves tracking by backpropagating returns through short imagined
rollouts.
HuScale follows this model-based approach, freezing the shared tracker across
demonstration collection, {KeypointVLA training}, and execution
with a shared spatial interface.

\section{Method}
\label{sec:method}

\begin{figure*}[!t]
    \centering
    \includegraphics[width=\linewidth]
        {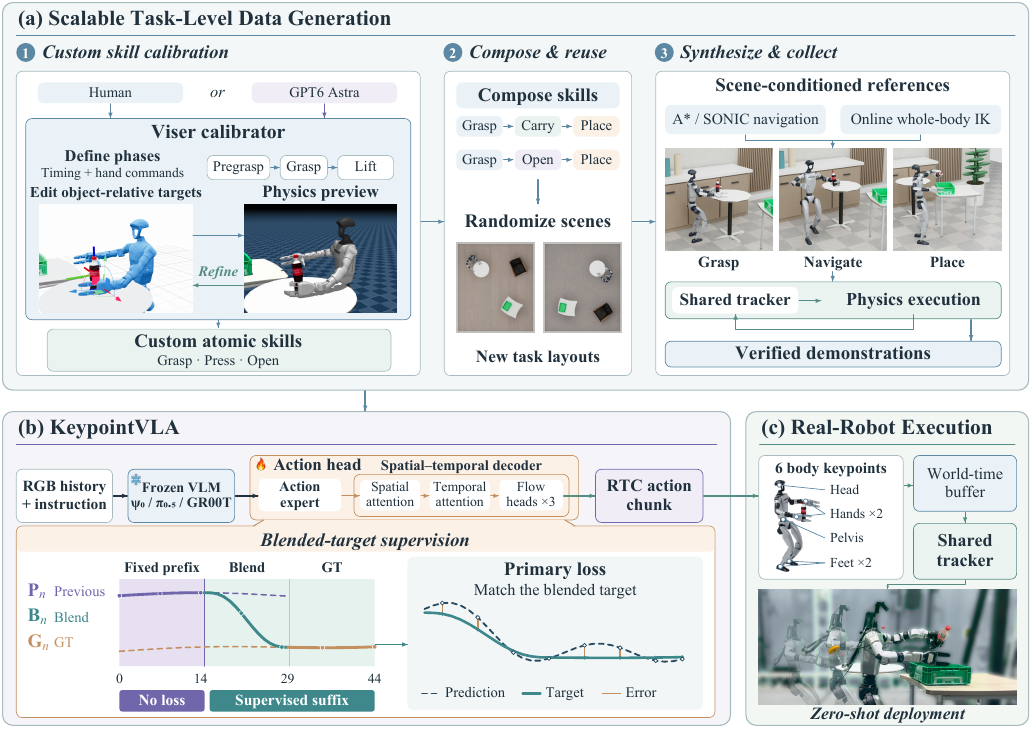}

    \caption{\textbf{HuScale framework.}
(a) Calibrate atomic skills, compose randomized tasks, and collect demonstrations.
(b) KeypointVLA with blended-target supervision: previous targets $\mathbf P_n$
and ground truth (GT) $\mathbf G_n$ form $\mathbf B_n$; primary loss matches
the predicted suffix to this target.
(c) Six-keypoint execution through the shared tracker.}
    \label{fig:pipeline}
\vspace{-2pt}
\end{figure*}

{We first describe how atomic skill calibration, task composition, and
scene-conditioned execution produce demonstrations (Fig.~\ref{fig:pipeline}(a)).
We then introduce KeypointVLA learning and closed-loop execution
(Fig.~\ref{fig:pipeline}(b,c)), followed by demonstration reuse,
visual augmentation, and real-robot deployment.}

\subsection{Task-Level Data Generation}
\label{sec:task_generation}

\paragraph{Atomic skills and interactive calibration.}
{An atomic skill is a reusable object-centric interaction,
such as opening a drawer. It specifies participating entities,
approach stances, ordered interaction phases, and success criteria.
Each phase stores object-relative body-keypoint targets, timing, and
hand commands. A task $\mathcal T=(\ell,[\sigma_1,\ldots,\sigma_M])$
combines an instruction $\ell$ with $M$ atomic skills.
In step 1 of Fig.~\ref{fig:pipeline}(a), humans or GPT6 Astra calibrate
custom atomic skills through the same Viser calibrator.
Humans edit targets and phases interactively; GPT6 Astra uses tool calls
to perform the same operations.
Physics previews combine online whole-body IK with the shared tracker,
letting robot motion and object responses guide refinement.}

\paragraph{{Task instantiation and reference synthesis.}}
{Step 2 of Fig.~\ref{fig:pipeline}(a) composes compatible atomic skills
into different tasks and reuses them across randomized scene layouts.
Their object-relative targets move with the associated objects.
We sample robot, object, and fixture poses within scene-feasible ranges,
keeping each task's semantics and physical parameters fixed.
The compiler checks skill preconditions and propagates interaction
effects, then maps calibrated object-relative targets into world coordinates:}
{
\begin{equation}
    {}^{W}\!K_{i,t}^{\mathrm{ref}}
    ={}^{W}\!T_{o,t}\,{}^{o}\!K_{i,t}^{\mathrm{ref}},
    \qquad T,K\in\mathrm{SE}(3).
\label{eq:object_to_world_reference}
\end{equation}
}
{The two reference-generation branches in step 3 of
Fig.~\ref{fig:pipeline}(a) handle navigation and manipulation.
A* routes are converted into locomotion references by a pretrained SONIC~\cite{luo2026sonic}
prior. Online whole-body IK solves manipulation
references from interpolated keypoint targets and the current measured
robot pose, subject to joint and root bounds.
Targets are interpolated at 50\,Hz using quintic position profiles and
quaternion SLERP.
Both branches share a timeline with phase-aligned hand commands.
Regenerating interactions from the achieved state yields varied
trajectories without per-demonstration manual editing.}

\paragraph{Closed-loop execution and synchronized demonstrations.}
{The physics-execution block in Fig.~\ref{fig:pipeline}(a)
tracks these references and feeds live state back to reference generation.
In simulation, keypoint errors trigger IK replanning at selected phase
boundaries. We retain complete rollouts satisfying final outcomes and
persistent conditions from preceding interactions.}

{At each sampled time $t_n$, we pair the egocentric
observation $I_{t_n}$ and instruction $\ell$ with action chunk $\mathbf A_n$.
Labels are commanded keypoint and wrist/finger targets; achieved states
and low-level control traces are auxiliary records.}

\subsection{\texorpdfstring{{KeypointVLA with Whole-Body Tracking}}{KeypointVLA with Whole-Body Tracking}}
\label{sec:spatial_policy}

{Figure~\ref{fig:pipeline}(b) shows action prediction and its training
supervision; Fig.~\ref{fig:pipeline}(c) shows execution through the
shared six-keypoint interface.}

\paragraph{Whole-body spatial representation.}
{Each action chunk~\cite{zhao2023act,chi2023diffusionpolicy} $\mathbf A_n\in\mathbb R^{45\times62}$
contains 45 targets at 20\,ms intervals.
The six-keypoint interface in Fig.~\ref{fig:pipeline}(c) comprises the pelvis,
head, both hands, and both feet. Each target contains their positions
$\mathbf p^k\in\mathbb R^3$ and unit $xyzw$ quaternions
$\boldsymbol\rho^k\in\mathbb S^3$, plus six wrist and fourteen finger angles:}
{
\begin{equation}
    \mathbf a_{n,j}
    =\big[(\mathbf p^k_{n,j},\boldsymbol\rho^k_{n,j})_{k=1}^{6},
          \mathbf w_{n,j},\mathbf f_{n,j}\big]\in\mathbb R^{62}.
\label{eq:spatial_action_row}
\end{equation}
}
{Each action chunk uses a camera XY/yaw frame fixed at
prediction time, with the camera's ground projection as origin and its
horizontal viewing direction as yaw.}

\paragraph{{History-conditioned action prediction.}}
{Every 0.3\,s, KeypointVLA predicts
$\widehat{\mathbf A}_n\sim\pi_\theta(\cdot\mid\mathcal O_n,\ell,\mathbf S_n,\mathbf P_n)$.
Here, $\mathcal O_n$ is image history, $\mathbf S_n\in\mathbb R^{10\times62}$
is measured-state history, and $\mathbf P_n$ contains the previous chunk's
last 30 time-aligned targets. Its first 15 valid targets form a fixed prefix
for real-time chunking (RTC)~\cite{black2025rtc,black2025trainingrtc}. Previous keypoint poses are transported into the current camera
frame; wrist and finger angles remain unchanged.}

\paragraph{{Spatial--temporal action decoder.}}
{Each frozen VLM feeds a common GR00T action-expert
architecture~\cite{bjorck2025grootn1,peebles2023dit} with learned context adapters and spatial conditioning;
action-head weights are trained separately for each backbone.
The spatial--temporal decoder in Fig.~\ref{fig:pipeline}(b) expands action-expert features
into six tokens per time step, applying spatial attention across keypoints
and temporal attention along each trajectory.
Three flow-matching heads~\cite{lipman2023flowmatching} predict flow
velocities for positions, orientations, and wrist/finger angles.
}

\paragraph{{Blended-target supervision.}}
{Training with ground-truth (GT) previous chunks provides no exposure
to model errors~\cite{ross2011dagger,chen2024diffusionforcing}. Previous chunk injection adds $\mathbf P_n$ predicted
by the training model from preceding samples, teaching correction toward GT.
However, direct GT supervision risks an abrupt transition from the fixed RTC prefix
to $\mathbf G_n$. We therefore use a detached blended target trajectory
$\mathbf B_n$ (Fig.~\ref{fig:pipeline}(b)).
For position and joint channels,}
{
\begin{equation}
    \mathbf B_{n,j}
    =w_j\mathbf P_{n,j}+(1-w_j)\mathbf G_{n,j},
\label{eq:blended_spatial_target}
\end{equation}
}
{Orientations use $\mathrm{SO}(3)$ interpolation with the same weights.
When the previous chunks are model-generated, valid targets in the first 15 steps
are fixed to previous chunks, without direct loss. The next 15 blend toward GT with $w_j$ decreasing
to zero, just as shown in Fig.~\ref{fig:pipeline}(b); the final 15 use GT.
For GT previous chunks, the suffix uses GT directly.}

{The primary loss compares predicted suffix targets with the blended target
$\mathbf B_n$.
It combines position, relative-displacement,
orientation, wrist/finger-angle, and rotational-flow errors.}

{The temporal loss supervises motion derivatives against the same blended target:}
{
\begin{equation}
    L_{\mathrm{temporal}}
    =\sum_{g\in\{p,R\}}\sum_{d=1}^{2}
      \lambda_{g,d}L_{g,d}
      +\lambda_{\mathrm{self}}\mathcal R_{\mathrm{acc}},
\label{eq:temporal_reference_loss}
\end{equation}
}{where $L_{g,d}$ matches first- or second-order
position/rotation derivatives to the blended target on valid stencils.
$\mathcal R_{\mathrm{acc}}$ regularizes position and wrist accelerations.
Fixed prefix targets receive no direct loss but supply detached boundary
conditions for suffix derivatives.}

{The total training objective is}
\begin{equation}
    L=L_{\mathrm{primary}}
    +\alpha L_{\mathrm{temporal}},
\label{eq:spatial_training_objective}
\end{equation}
{where adaptive $\alpha$ limits the temporal term's contribution to the total loss.}

\paragraph{{Learning the shared whole-body tracker.}}
{We train the tracker independently of KeypointVLA using a
SuperMimic-style objective~\cite{liao2026supermimic} and freeze it for
demonstration collection and deployment. 
The tracker follows the predicted body-keypoint targets
in closed loop, while wrist and finger targets use synchronized PD control.}

\subsection{\texorpdfstring{{Data}}{Data} Reuse and Real-World Extension}
\label{sec:data_reuse}

\paragraph{SONIC-compatible reuse.}
{Following GRAIL~\cite{xie2026grail}, we optionally train
task-specific RL motion adaptors to re-execute collected references with
native SONIC~\cite{luo2026sonic}. Only successful re-executions are retained
for this alternative interface, evaluated in Sec.~\ref{sec:interface_comparison}.}

\paragraph{Replay-based visual augmentation.}
{We rerender verified demonstrations under varied visual
conditions~\cite{tobin2017domainrandomization} and pair the resulting images with the original task instructions
and action labels at matching control steps. This expands visual coverage
without adding demonstrations.}

\paragraph{{Real-robot collection.}}
{On physical robots, we use the same calibrated atomic
skills and reference generation procedure. We record observations and
commanded targets synchronously and retain successful rollouts. Between
phases, the robot waits for the online IK solution, while the simulation
clock pauses during each solve.}

\section{Experiments}
\label{sec:experiments}

We evaluate task customization and collection efficiency, six-task
KeypointVLA performance, previous chunk injection and closed-loop
execution, comparison with SONIC, and real-world collection and deployment.

\subsection{Experimental Setup}
\label{sec:exp_setup}

\paragraph{Robot and task suite.}
We use a Unitree G1 in MuJoCo~\cite{todorov2012mujoco} for six tasks: Pick-and-Lift (T1),
Pick-and-Place (T2), Pedal Disposal (T3), Bedside Cabinet (T4),
Drawer Storage (T5), and Two-Object Transport (T6).
Collection and KeypointVLA use a shared whole-body tracker, frozen during both stages.

\paragraph{Scene randomization.}
Scene layouts combine shared robot--scene transforms with local pose
perturbations. Objects are sampled within feasible regions on supporting
surfaces; materials, backgrounds, and lighting are varied.
Only layouts satisfying support, clearance, room-boundary, and
initial-visibility constraints are executed. Detailed randomization and implementation settings are provided in the
supplementary material.

\paragraph{Data and training.}
The six-task corpus contains 12,000 accepted demonstrations, 2,000 per task.
Each backbone ($\psi_0$, $\pi_{0.5}$~\cite{black2025pi05}, and GR00T) is trained jointly on all
six tasks using frozen VLM weights and separately trained action-head
weights. Visual re-renders do not add demonstrations.

\paragraph{Evaluation protocol.}
The six-task benchmark uses 30 held-out scenes per task
(25 within training ranges, five beyond), with three robot starts each
(90 trials). 
Placement requires at least 80\% object volume inside the target;
drawer closure allows at most 25\% remaining opening.

\subsection{Task Expansion and Collection Efficiency}
\label{sec:task_expansion}

\paragraph{Task customization and atomic skill reuse.}
Table~\ref{tab:task_authoring_time} separates calibration from
post-calibration collection. HuScale's recorded session averages
7.3 minutes per atomic skill.
With calibration costs comparable to other methods, HuScale supports
full task customization with the shortest
50-demo collection time among the compared pipelines, even up to a $74\times$ speedup
over end-to-end GPT6 Astra. Once calibrated, these skills can generate
diverse demonstrations of the same task across randomized scenes without
further manual adjustment. What's more,
seven calibrated atomic skills can be composed into tens of
tasks, of which T1--T6 are six representative examples.

\begin{figure}[!htb]
\centering
\includegraphics[width=\linewidth]{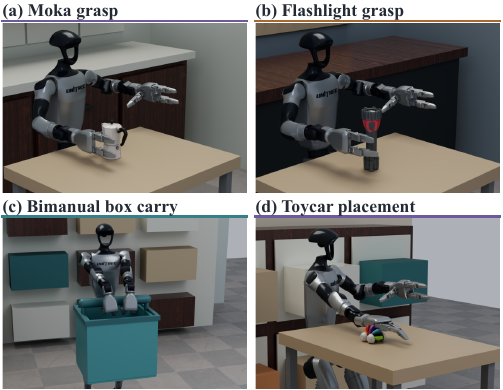}
\caption{\textbf{GPT6 Astra calibration examples.}
Isaac Sim renders across assets and scenes.}
\label{fig:astra_calibration}
\vspace{-4pt}
\end{figure}

\paragraph{GPT6 Astra batch calibration.}
GPT6 Astra calibrated \textbf{41 distinct atomic skills across 37 assets},
counting unique object--interaction pairs with at least one accepted
physical preview. The recorded interval from the first physical-preview
launch to the saved batch report lasted 1.87\,h.
Figure~\ref{fig:astra_calibration} illustrates how GPT6 Astra calibration
extends the reusable atomic skill library across diverse assets.

\begin{table}[!htb]
\centering\small\setlength{\tabcolsep}{3pt}
\renewcommand{\arraystretch}{1.08}
\caption{
Calibration and 50-demo collection time (min).
$-$: unavailable; $/$: no calibration step.
}
\label{tab:task_authoring_time}
\begin{tabularx}{\linewidth}{@{}>{\raggedright\arraybackslash}p{0.47\linewidth}*{2}{>{\centering\arraybackslash}X}@{}}
\toprule
Method / setup & \makecell{Calibration\\time} & \makecell{50-demo\\time} \\
\midrule
LEGS~\cite{kim2026legs} & $<15$ & $30$ \\
SIMPLE (MP)~\cite{wei2026simple} & 6.6 & $125$ \\
Zero-HLM~\cite{yang2026zerohlm} & - & 47.6 \\
\midrule
HuScale (interactive) & 7.3 & $4.4$ \\
HuScale (reuse) & 0 & $4.4$ \\
GPT6 Astra (end-to-end) & / & 324.9 \\
\bottomrule
\end{tabularx}
\vspace{-2pt}
\end{table}

\begin{figure}[!htb]
\centering
\includegraphics[width=\linewidth]{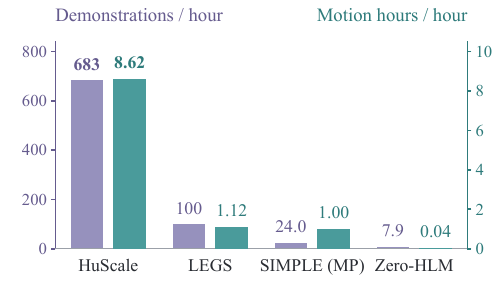}
\caption{\textbf{Data generation throughput per GPU.}
Purple: trajectories/hour; teal: motion hours/hour.}
\label{fig:collection_throughput}
\vspace{-2pt}
\end{figure}

\paragraph{Data generation throughput.}
On an NVIDIA RTX~4090 (48\,GB), HuScale reaches
\textbf{683 trajectories and 8.62 motion hours per wall-clock hour}.
Figure~\ref{fig:collection_throughput} compares these rates with published
automated loco-manipulation pipelines. Both rates support the scalability.

\subsection{KeypointVLA Across Six Tasks}
\label{sec:vla_scaling}

\begin{figure}[!b]
\centering
\includegraphics[width=\linewidth]{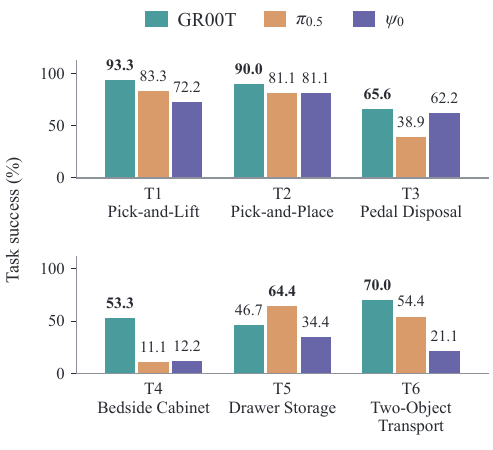}
\caption{\textbf{KeypointVLA success across six tasks.}}
\label{fig:data_scaling}
\vspace{-2pt}
\end{figure}

\paragraph{Backbone performance.}
Figure~\ref{fig:data_scaling} shows successful loco-manipulation across
six tasks with each of the three pretrained VLM backbones.
GR00T performs best on most tasks, while $\pi_{0.5}$ leads on Drawer
Storage (T5). Together, these results demonstrate the utility of HuScale's
generated demonstrations and the effectiveness of KeypointVLA's spatial
action design across different backbone families and interaction types.

\subsection{Previous Chunk Injection}
\label{sec:feedback_ablation}

\begin{figure}[!htb]
\centering
\includegraphics[trim=0 14bp 0 0,clip,width=\linewidth]{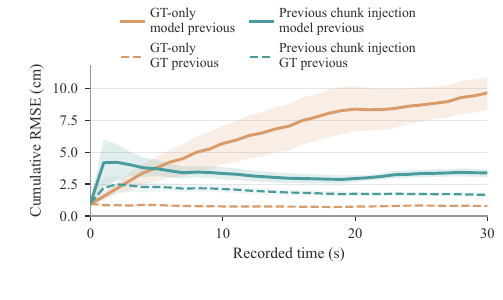}
\caption{\textbf{Autoregressive prediction error.}
Solid/dashed: causal model/GT previous chunks.
Shading: pointwise 95\% episode-bootstrap intervals.}
\label{fig:feedback_error_curve}
\vspace{-2pt}
\end{figure}

\paragraph{Recorded-observation protocol.}
We compare two historical GR00T checkpoints trained without and with
previous chunk injection. The former uses GT previous chunks during
training. For each checkpoint, we fix recorded RGB/state observations and
supply either GT previous chunks or causal model predictions.
Figure~\ref{fig:feedback_error_curve} reports cumulative end-effector
position RMSE over 50 training episodes.
Although the GT-only checkpoint achieves lower one-step error with
ground-truth previous chunks, its errors accumulate during autoregressive
rollout. With previous chunk injection, the checkpoint has slightly higher one-step
error but suppresses this accumulation, yielding substantially more stable
long-horizon predictions.

\begin{figure}[!htb]
\centering
\includegraphics[width=\linewidth]{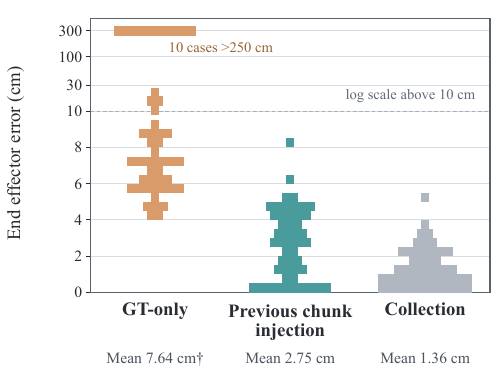}
\caption{\textbf{End effector error distributions.}
Width: episode frequency. Errors above 250\,cm are pooled;
the axis is logarithmic above 10\,cm.
$\dagger$GT-only mean excludes ten pooled outliers.}
\label{fig:feedback_gap_distribution}
\vspace{-2pt}
\end{figure}

\paragraph{Physical execution.}
Both checkpoints execute in simulation once for 90\,s in 50
scenes from collection data, simulating asynchronous autoregressive inference
under real-world deployment conditions. With previous chunk injection,
end effector errors, measured as palm-to-bottle surface distances,
cluster closer to the collection reference
(Fig.~\ref{fig:feedback_gap_distribution}), indicating more reliable
interaction localization during physical execution.

\subsection{KeypointVLA Execution Comparison}
\label{sec:interface_comparison}
\label{sec:policy_analysis}

\begin{figure}[!b]
\centering
\includegraphics[trim=0 9bp 0 0,clip,width=\linewidth]{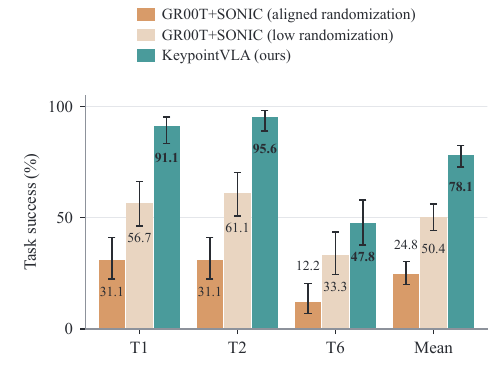}
\caption{\textbf{KeypointVLA versus GR00T+SONIC.}
Error bars show 95\% Wilson intervals.}
\label{fig:interface_comparison}
\vspace{-2pt}
\end{figure}

\begin{figure*}[!t]
\centering
\includegraphics[width=0.95\linewidth]{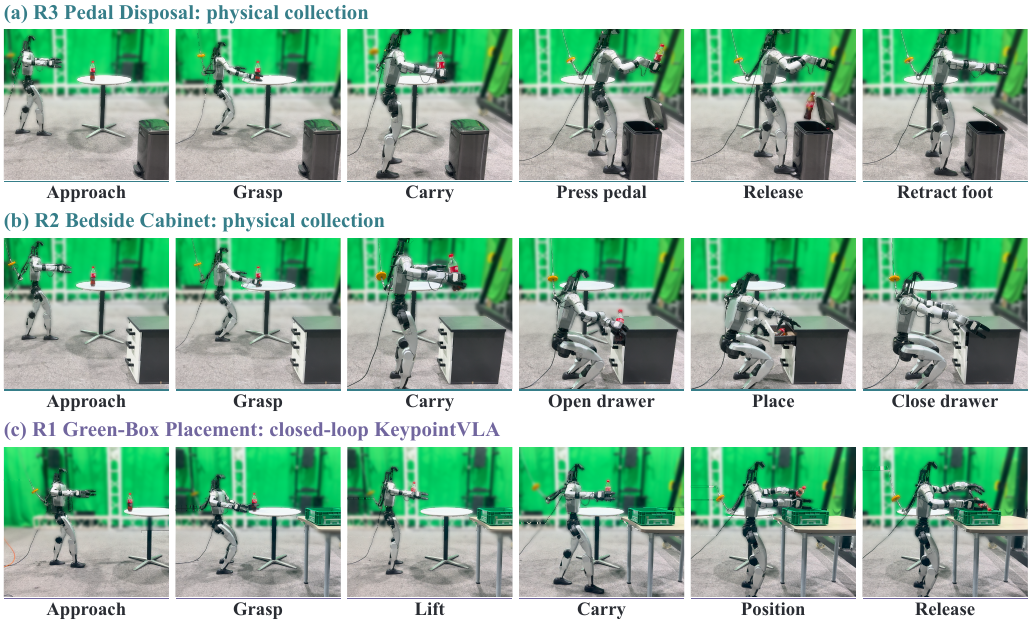}
\vspace{4pt}
\caption{\textbf{Real-world collection (a,b) and zero-shot KeypointVLA deployment (c).}}
\label{fig:real_world}
\vspace{4pt}
\end{figure*}

\paragraph{Comparison with SONIC-based execution.}
Following GRAIL~\cite{xie2026grail}, we train RL adaptors with
task-specific rewards to obtain SONIC-compatible~\cite{luo2026sonic}
re-executions of HuScale references.
We adapt its released pick-and-place reward implementation and restrict
the comparison to T1, T2, and T6, which are compatible with this setup.
We convert 200 demonstrations per task and train both policies with this
budget, evaluating them on the same 90 trials per task under aligned
randomization. We also regenerate 200 demonstrations per task under lower
randomization to train a separate native GR00T+SONIC model, evaluated
under the narrower setting.

\paragraph{Closed-loop task success.}
Figure~\ref{fig:interface_comparison} shows that KeypointVLA outperforms
GR00T+SONIC on all three tasks under aligned randomization, achieving
78.1\% versus 24.8\% overall success.
Reducing training and evaluation randomization improves GR00T+SONIC to
50.4\%, confirming that it can learn effective policies in simpler
settings; KeypointVLA remains ahead despite the broader scene distribution.
The advantage also extends to multi-stage Two-Object Transport (T6),
supporting KeypointVLA's ability to sustain closed-loop task completion
across varied scenes and composed interactions.

\subsection{Real-World Collection and Policy Execution}
\label{sec:real_world}

A replacement G1 head carries a ZED~2i stereo camera for a wider field of
view, supplying left and right RGB images without depth maps to the policy.
We distinguish collector-driven demonstrations from learned policy
execution on Green-Box Placement (R1), Bedside Cabinet (R2), and
Pedal Disposal (R3).

\paragraph{Physical demonstration collection.}
The collector uses calibrated atomic skills and object poses with online
IK and the shared tracker. We use success rate and mean execution time
as collection metrics (Table~\ref{tab:real_robot_collection}).
Figure~\ref{fig:real_world}(a,b) illustrates R3 and R2 collection,
including approach and multi-stage interactions, supporting real-robot
collection through the shared spatial interface.

\paragraph{Zero-shot deployment from synthetic demonstrations.}
We train GR00T-based KeypointVLA with 2,000 synthetic demonstrations per
task and material/lighting augmentation, without real-robot demonstrations
or real-data fine-tuning. The deployment protocol uses the same tasks and
metrics (Table~\ref{tab:real_robot_collection}).
Figure~\ref{fig:real_world}(c) shows qualitative closed-loop execution on R1.
These results support transferring synthetic demonstrations to physical
closed-loop execution.

\begin{table}[!htb]
\centering\small\setlength{\tabcolsep}{3pt}
\vspace{3pt}
\caption{Real-World Collection and Zero-Shot Deployment.}
\label{tab:real_robot_collection}
\vspace{4pt}
\begin{tabularx}{\linewidth}{@{}l*{4}{>{\centering\arraybackslash}X}@{}}
\toprule
\multirow{2}{*}[-.2em]{Task} & \multicolumn{2}{c}{Collection} & \multicolumn{2}{c}{Zero-shot deployment} \\
\cmidrule(lr){2-3}\cmidrule(lr){4-5}
 & Success & Time (s) & Success & Time (s) \\
\midrule
R1: Box & 9/10 & 58 & 8/10 & 39 \\
R2: Cabinet & 7/10 & 109 & 7/10 & 58 \\
R3: Bin & 5/10 & 71 & 4/10 & 47 \\
\bottomrule
\end{tabularx}
\vspace{4pt}
\end{table}

\paragraph{Failure analysis.}
In real-world VLA rollouts, visual errors can accumulate over long-horizon tasks and lead to inaccurate distance estimates. During sustained crouching or single-leg stance, joint actuator overheating can cause low-level tracking oscillations and compromise balance. Both can lead to task failure.

\section{Conclusion}
\label{sec:conclusion}

We presented HuScale, a pipeline for scalable task-level data generation
and humanoid loco-manipulation.
By composing reusable atomic skills calibrated by humans or GPT6 Astra,
HuScale generates physically feasible demonstrations without
per-trajectory editing.
KeypointVLA learns six-keypoint trajectories with previous chunk
injection and blended-target supervision, using the shared whole-body
tracker for closed-loop execution.
Experiments across six tasks and three VLA backbones support the utility
of the generated data, while G1 demonstrations illustrate real-robot collection and zero-shot deployment.
Future work will extend the data generation pipeline to support more diverse and complex tasks, such as dexterous in-hand manipulation.

\section*{LLM Usage Statement}
GPT6 Astra was used to calibrate atomic skills
(Sec.~\ref{sec:task_expansion}), including proposing and refining
interaction targets using physics-based preview feedback.
OpenAI Codex assisted with manuscript drafting and revision,
literature searches and reference checks, implementation and debugging
of research and evaluation code, analysis of experiment logs,
and preparation of figures and tables.
The authors remain responsible for the research design, validation,
interpretation, and all submitted content.


{
    \small
    \setlength{\bibsep}{0pt}
    \bibliographystyle{ieeenat_fullname}
    \bibliography{main}
}


\end{document}